\documentclass{easychair}

\usepackage[T1]{fontenc}
\usepackage[utf8]{inputenc}
\usepackage{graphicx}
\usepackage{amsmath}
\usepackage{amssymb}
\usepackage[ruled,linesnumbered,vlined]{algorithm2e}

\usepackage{xcolor}
\definecolor{normalbg}{HTML}{EDF2FC}
\definecolor{normalfg}{HTML}{012528}
\definecolor{examplebg}{HTML}{648381}
\definecolor{alertbg}{HTML}{FAC05E}
\definecolor{myblue}{HTML}{DA4167}

\usepackage{tabularx}
\usepackage{tabu}
\usepackage{url}
\usepackage{hyperref}
\usepackage{caption}
\usepackage{xspace}

\usepackage{pifont}
\newcommand{\cmark}{{\color{myblue}\ding{51}}}
\newcommand{\xmark}{\ding{55}}

\usepackage{tikz}
\usetikzlibrary{positioning}
\usetikzlibrary{shapes,arrows}
\usetikzlibrary{arrows.meta}
\usetikzlibrary{calc}
\usepackage[export]{adjustbox}

\usepackage{mathtools}

\usepackage{soul}
\definecolor{amethyst}{rgb}{0.6, 0.4, 0.8}

\newcommand{\satviz}{\textsf{SATViz}\xspace}

\title{\satviz: Real-Time Visualization of Clausal Proofs}

\author{
Tim~Holzenkamp
\and
Kevin~Kuryshev
\and
Thomas~Oltmann
\and
Lucas~Wäldele
\and
Johann~Zuber
\and
Tobias~Heuer\thanks{Project Supervision}
\and
Ashlin~Iser\thanks{Project Definition and Supervision}
}

\institute {
Algorithm Engineering, Institute of Theoretical Informatics, KIT-Department of Informatics, Karlsruhe Institute of Technology (KIT), Germany
}

\authorrunning{Holzenkamp et al.}
\titlerunning{\satviz}

\begin{document}
\maketitle
\begin{abstract}
Visual layouts of graphs representing SAT instances can highlight the community structure of SAT instances.
The community structure of SAT instances has been associated with both instance hardness and known clause quality heuristics.
Our tool \satviz visualizes CNF formulas using the variable interaction graph and a force-directed layout algorithm.
With \satviz, clause proofs can be animated to continuously highlight variables that occur in a moving window of recently learned clauses.
If needed, \satviz can also create new layouts of the variable interaction graph with the adjusted edge weights.
In this paper, we describe the structure and feature set of \satviz.
We also present some interesting visualizations created with \satviz.
\end{abstract}

\section{Introduction}

Visual representations of algorithms can improve our understanding of algorithmic concepts and the nature of the underlying problems.
Visualizations of the Conflict-Driven Clause Learning algorithm (CDCL) are of high interest, as this is currently the most successful algorithm for solving the propositional satisfiability problem (SAT problem).
Despite the complexity of the SAT problem, implementations of CDCL in so-called SAT solvers are successfully used in industrial practice, e.g., in software verification~\cite{Buning:2020:QPRverify}, hardware verification~\cite{Kaufmann:2021:AMulet}, product configuration~\cite{Werner:2021:ValDom}, or planning~\cite{Schreiber:2021:Lilotane}.

For satisfiable instances, SAT solvers output a variable assignment which serves as a certificate of satisfiability.
Such a certificate can be checked in polynomial time and hence SAT~$\in$~NP.
However, modern SAT solvers also output certificates of unsatisfiability -- a proof given in the DRAT format~\cite{drat-trim}.
Such a proof can be checked efficiently by a procedure with runtime polynomial in the proof length. 
The output of verifiable proofs increases the trust in the correctness of SAT algorithms.
Trust is crucial in scenarios where SAT solvers, e.g., verify the absence of bugs in safety-critical software.

Famous proofs of unsatisfiability that solved previously open mathematical problems, e.g., the Boolean Pythagorean triples problem, have been generated by SAT solvers~\cite{Heule:2018:Schur,Heule:2016:Pyth}. 
The explainability of such automatically generated proofs is of great interest. 
This is mainly due to their sheer size, e.g., the total size of the proof of the Boolean Pythagorean triples problem is 200 TB. 

The source of such generated proofs are the clauses learned by a CDCL SAT solver. 
In CDCL, learned clauses, i.e., small proofs, also help to accelerate search-space navigation. 
The structure of the resulting reasoning is hard to grasp as millions of clauses are involved. 
The average instance of the SAT competition~2021 benchmark contains $2.3$ millions of clauses. 
When running \textsf{Kissat}~\cite{kissat} on these instances we observe an average of 8676 (with a median of 4040) learned clauses per second. 

To this end, visual representations illustrating learned clauses during the execution of SAT solvers
can bridge the gap to gain a better understanding of the structure of automated reasoning and resolution-space navigation. 
In this paper, we present a tool which creates real-time visualizations of proof-sequences as they are created by a given SAT solver or proof file.

\section{Background}

Instances of the SAT problem are given in \emph{Conjunctive Normal Form} (CNF) and are represented by a \emph{set of clauses}.
Each clause is a is a \emph{set of literals} and represents the disjunction of those literals.
Each literal is a negated or non-negated Boolean variable.

CDCL combines search (cf. DPLL algorithm) and resolution (cf. saturation algorithm)~\cite{sat-handbook}. 
CDCL solvers maintain and extend a current partial assignment which is created from repeated decision and subsequent Boolean constraint propagation. 
On conflict, CDCL infers a new clause by resolution of the conflict-reason clauses which are determined by analyzing the implication graph of the current partial assignment~\cite{Feng:2020:AllUIP}. 
If the empty clauses can be inferred, then the set of learned clauses forms a testable proof of the unsatisfiability of an instance. 
Clause learning is also an important means of narrowing down the search space of satisfiable instances, e.g., to escape from unsatisfiable regions of the search space early on.
As the relevance of a learned clause for the overall search process can not be known in advance, clause learning and forgetting is controlled by heuristics~\cite{Oh:2015:three-tier}. 

A SAT formula in conjunctive normal form (CNF) can be represented as a hypergraph.
Hypergraphs are generalizations of regular graphs in which each \emph{hyperedge} can connect more than two nodes.  In this model, each variable corresponds to a node and each clause is modeled as a hyperedge.
While there is little work on visualizing hypergraphs, we transform the hypergraph to a regular graph.
The variant in which each hyperedge is replaced by a clique is known as the \emph{Variable Interaction Graph} (VIG) in the SAT domain (cf.~Section~\ref{sec:reduction}).

\section{Related Work}

\begin{table}[t]
\centering
\begin{tabular}{l|ccccc}
							& DPvis & 3Dvis & iSAT & SATGraf & \satviz \\
\hline
Interactive					& \cmark & \xmark & \cmark & \cmark & \cmark \\
Multiple Reduction Types	& \xmark & \xmark & \cmark & \cmark & \cmark \\
3D Layout					& \xmark & \cmark & \xmark & \xmark & \xmark \\
Real-time Animations		& \cmark & \xmark & \xmark & \cmark & \cmark \\
Graph Contractions			& \xmark & \xmark & \xmark & \xmark & \cmark \\
IPASIR Interface	 		& \xmark & \xmark & \xmark & \xmark & \cmark \\
DRAT Interface	 			& \xmark & \xmark & \xmark & \xmark & \cmark \\
\end{tabular}
\caption{Feature Coverage of CNF Formula Visualizers}
\label{tab:comparison}
\end{table}

In the area of SAT solving, the tools \textsf{DPvis} and \textsf{3Dvis} by Sinz have been used to visualize the structure of SAT instances based on the force-directed layout of their graph representations~\cite{Sinz:2007:DPvis}.
\textsf{DPvis} can also visualize formula evolution by simplifying according to the branching and propagation steps in runs of the integrated SAT solver Minisat~\cite{minisat}.
Visualizations of \textsf{3Dvis} appear in Knuth's ``The Art of Computer Programming'' Vol. 4.6 on ``Satisfiability'' (the author's favorite pages)~\cite{Knuth:2015:TaocpSat}.
Recent SAT competitions use the intriguing visualizations of 3Dvis as their logos.\footnote{\url{https://satcompetition.github.io/2022/logo2022-large.png}}

\textsf{iSAT} is a tool for instrumentation and interactive control of SAT solvers.
\textsf{iSAT} facilitates the analysis of intermediate solver states.
For external visualization, \textsf{iSAT} can export several types of graph representations of formulas to files~\cite{Orbe:2012:iSAT}.
The tool \textsf{SATGraf} by Newsham et al. visualizes community structure of SAT formulas.
SATGraf can also display statistics and solving progress of specially instrumented SAT solvers running on small SAT instances~\cite{Newsham:2015:SATGraf}.

Table~\ref{tab:comparison} shows a comparison of the features covered by \satviz and related tools.
Except for the more performance-critical \textsf{3Dvis}, all tools considered offer some form of interactive control to reinitialize the visualization with the current solver state.
While \textsf{DPvis} and \textsf{3Dvis} are limited to the use of the VIG, the other tools offer additional options and some form of extensibility.
\textsf{3Dvis} is the only tool that can generate three-dimensional layouts of SAT instances.

Real-time animations are supported by \textsf{DPvis}, \textsf{SATGraf} and \satviz, but each of them highlights a different aspect.
\textsf{DPvis} focuses on search progress via unit resolution, \textsf{SATGraf} focuses on the survival of the pre-computed communities, and \satviz highlights recently learned clauses.
Both \textsf{SATGraf} and \satviz can relayout an instance which has been modified by learned and forgotten clauses.

However, \satviz is the only tool that facilitates the handling of large industrial SAT instances in real-time.
This is possible since \satviz offers graph contractions for reducing the number of edges while preserving the graph structure.
Like this, even large SAT instances can be visualized efficiently and  also become less cluttered.

\satviz instruments SAT solvers with standard methods of the IPASIR~\cite{ipasir} or can moreover replay proofs given in the DRAT format~\cite{drat-trim}.
In contrast to other tools, \satviz can visualize the course of any SAT solver which implements the IPASIR or exports DRAT proofs.

\section{The Tool \satviz}

With our tool \satviz, we can transform CNF formulas to graphs and visualize them using a graph drawing algorithm.
A SAT solver running on the local or a remote machine can then connect
and send learned clauses to the application, which are then highlighted in the visual graph representation.
Thereby, the color of a node represents how frequently the corresponding variable appeared in learned clauses (similar to a heat-map).
Through this visualization we can better analyze the working of CDCL algorithms and better understand the structure of different problem instances. 
Code and instructions for using \satviz can be found on GitHub.\footnote{\url{https://github.com/satviz/satviz}}

\subsection{Architecture}

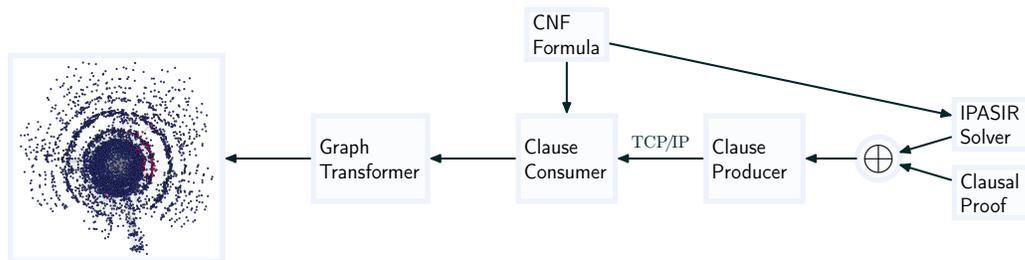
\begin{figure}
\begin{adjustbox}{width=\textwidth}
\begin{tikzpicture}[-{Latex[round]}, line width=1.2pt, normalfg, node distance=5em]
\tikzset{default/.style={black, draw=normalbg, line width=3pt, fill=normalbg!20, minimum size=0.5cm, font=\relsize{1.5}\sffamily}}

\node at (current page.west) (left) {};
\node at (current page.east) (right) {};
\node at (current page.center) (center) {};

\node[default, text width=5em, minimum height=5em] (cons) at (current page.center) { Clause Consumer };
\node[default, right=of cons, text width=5em, minimum height=5em] (prod) { Clause Producer };
\node[default, circle, right=of prod, inner sep=0, outer sep=0, xshift=-2em] (oplus) {\Huge $\oplus$};
\node[default, right=of oplus, text width=4em, minimum width=4em, yshift=+2em, xshift=-2em] (ipasir) { IPASIR Solver };
\node[default, right=of oplus, text width=4em, minimum width=4em, yshift=-2em, xshift=-2em] (drat) { Clausal Proof };
\node[default, above=of cons, text width=4em, minimum width=4em, yshift=-2em] (cnf) { CNF Formula };

\draw (cnf) -- (ipasir) {};
\draw (cnf) -- (cons) {};

\draw (drat) -- (oplus) {};
\draw (ipasir) -- (oplus) {};
\draw (oplus) -- (prod) {};
\draw (prod) -- node [above] {TCP\!\textbf{/}\!IP} (cons);

\node[default, left=of cons, text width=6em, minimum height=5em] (trans) { Graph\\ Transformer };
\node[default, left=of trans] (layout) {\includegraphics[width=10em, trim={470 30 470 60px}, clip]{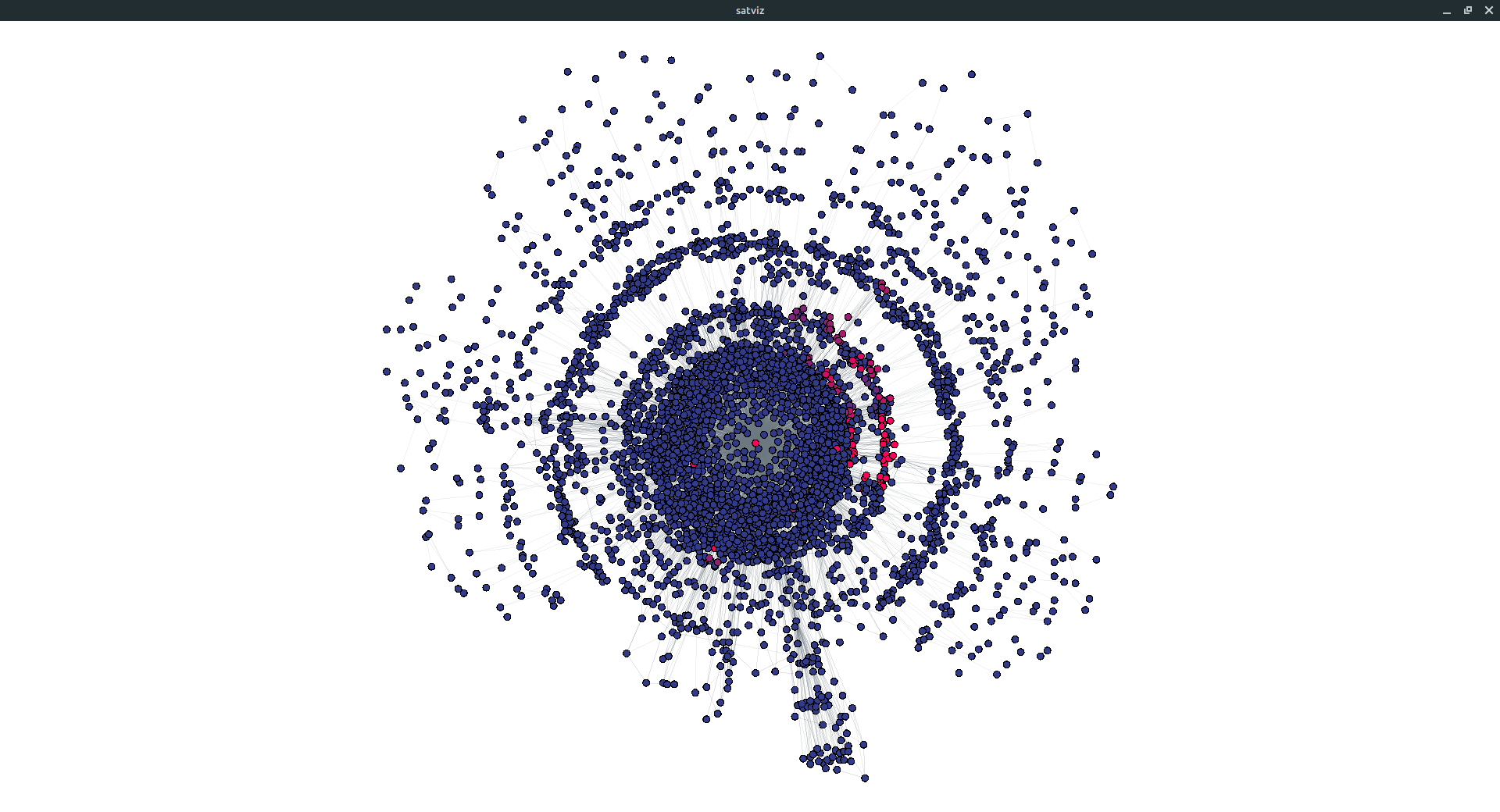}};

\draw (cons) -- (trans) {};
\draw (trans) -- (layout) {};

\end{tikzpicture}
\end{adjustbox}
\caption{The Architecture of \satviz}
\label{fig:arch}
\end{figure}

Figure~\ref{fig:arch} illustrates the core components of \satviz' architecture.
It can be divided into two central components: the \emph{clause consumer} and the \emph{clause producer}.
The clause consumer is responsible for visualizing SAT instances and learned clauses, while
the clause producer sends a stream of learned clauses to the clause consumer over the network.
This decoupling adds some flexibility to the architecture,
as learned clauses can be sent from different machines from any software implementing the clause producer protocol.

The clause producer of \satviz can read clauses from two types of sources: a file containing
a clausal proof or a shared library of an SAT solver implementing the IPASIR interface~\cite{ipasir}.
In the former, \satviz reads a clausal proof, e.g., in DRAT file format~\cite{drat-trim}, and streams it to the
clause consumer. In the latter, the clause producer starts a SAT solver initialized with a given CNF formula and
streams learned clauses to the clause consumer at the time they are exposed over the IPASIR interface
(intercepted in an \verb+ipasir_set_delete+ callback function).

The clause consumer buffers the incoming stream of learned clauses and
sends chunks of incoming clauses with a fixed frame rate to the \emph{graph transformer} component.
The graph transformer is responsible for visualizing learned clauses in the graph representation.
It adds new edges or updates their weight, and changes the colors of the corresponding nodes contained in the learned clauses.
This transformation is configurable and extensible.
The graph transformer configuration includes parameters to control the heat-map which highlights learned clauses (cf.~Section~\ref{sec:heatmap}),
different graph models to represent CNF formulas (cf.~Section~\ref{sec:reduction}), as well as options for reducing the size of larger instances
to improve the readability of the visualization and to speedup the performance of the animation (cf.~Section~\ref{sec:contraction}).

Given a graph, the layout algorithm uses force-directed placement to create the visual representation of the graph. 
We use the implementation provided by the Open Graph Drawing Framework (OGDF)~\cite{Chimani:2013:OGDF}.
The heat-map algorithm continuously updates node colors according to the incoming learned clauses.
The placement algorithm creates and uses an initial layout based on the original formula.
But on demand, it can recalculate a new layout based on the modified graph induced by the clauses which have been learned and forgotten so far.
The user interface also allows to pause, stop, rewind and replay the solver run and to step in at any time into the proof. 
For larger graph the magnification feature using the mouse wheel is very practical.

\satviz also supports to record the visualization including user interactions and exports this to a video file. 
\satviz can also operate in a headless mode and silently create videos from given proofs or SAT solver runs in the background. 
Some sample videos can be found in the dedicated YouTube playlist of Kuryshev.\footnote{\url{https://www.youtube.com/playlist?list=PLhFsx92qubLxMsFQ4lUlCS6vFmJvTbYJS}}

\subsection{Hypergraph Reduction}
\label{sec:reduction}

A CNF formula induces a hypergraph such that the variables correspond to nodes and each clause
forms a hyperedge spanning its variables.
Hypergraphs are hard to visualize directly, which is why reductions to
graph representations are used in practice~\cite{Sinz:2007:DPvis}.
The two most common representations are the \emph{clique expansion} and the \emph{bipartite graph representation}~\cite{BIPARTITE-GRAPH,BIPARTITE-GRAPH-2}.
The former adds a clique between all nodes contained in a hyperedge, while the latter additionally models the hyperedges as nodes and connects each
with their corresponding nodes. However, the clique expansion unnecessarily increases the size of our graph model in presence of large hyperedges, while
the bipartite graph representation adds much more nodes than necessary (the number of clauses is usually considerably larger than the number of variables).

We therefore present a third reduction to which we refer to as the \emph{ring reduction}. Here, we sort the node IDs of each hyperedge in increasing order
and connect consecutive nodes with an edge. Additionally, we add an edge between the two nodes with the lowest and highest ID in each hyperedge.
This representation combines the advantages of both standard representations as the number of nodes equals to the nodes in clique expansion, while
the number of edges equals to the edges in the bipartite graph representation.

In \satviz, we support the ring reduction and clique expansion as graph models.
Additionally, each hyperedge contributes an edge weight which is a configurable function inversely proportional to its own size.

%

\subsection{Graph Contraction}
\label{sec:contraction}

\begin{figure}[t]
\centering
\includegraphics[width=.5\linewidth, trim={450 0 450 40px}, clip]{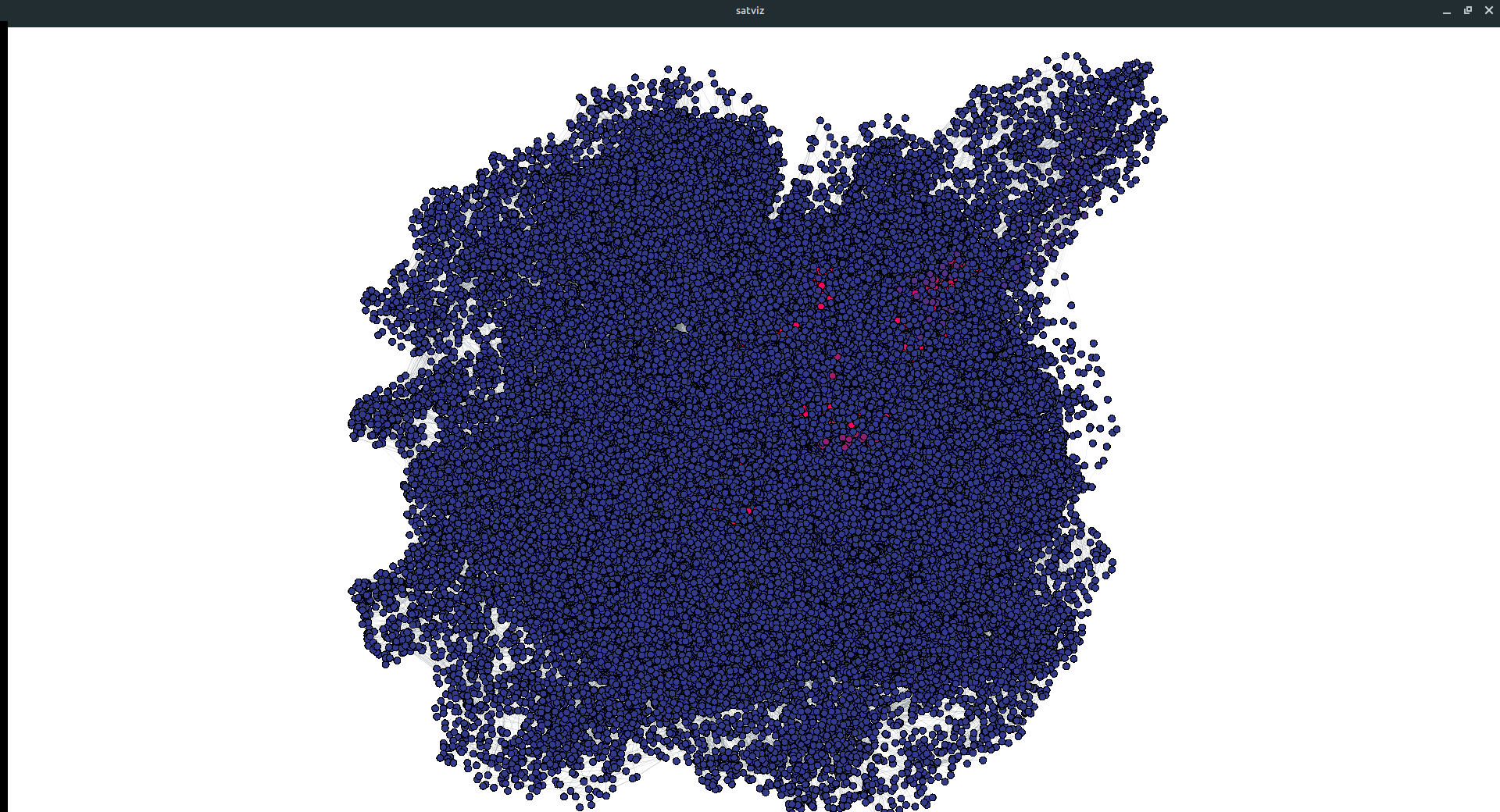}%
\includegraphics[width=.5\linewidth, trim={450 0 450 40px}, clip]{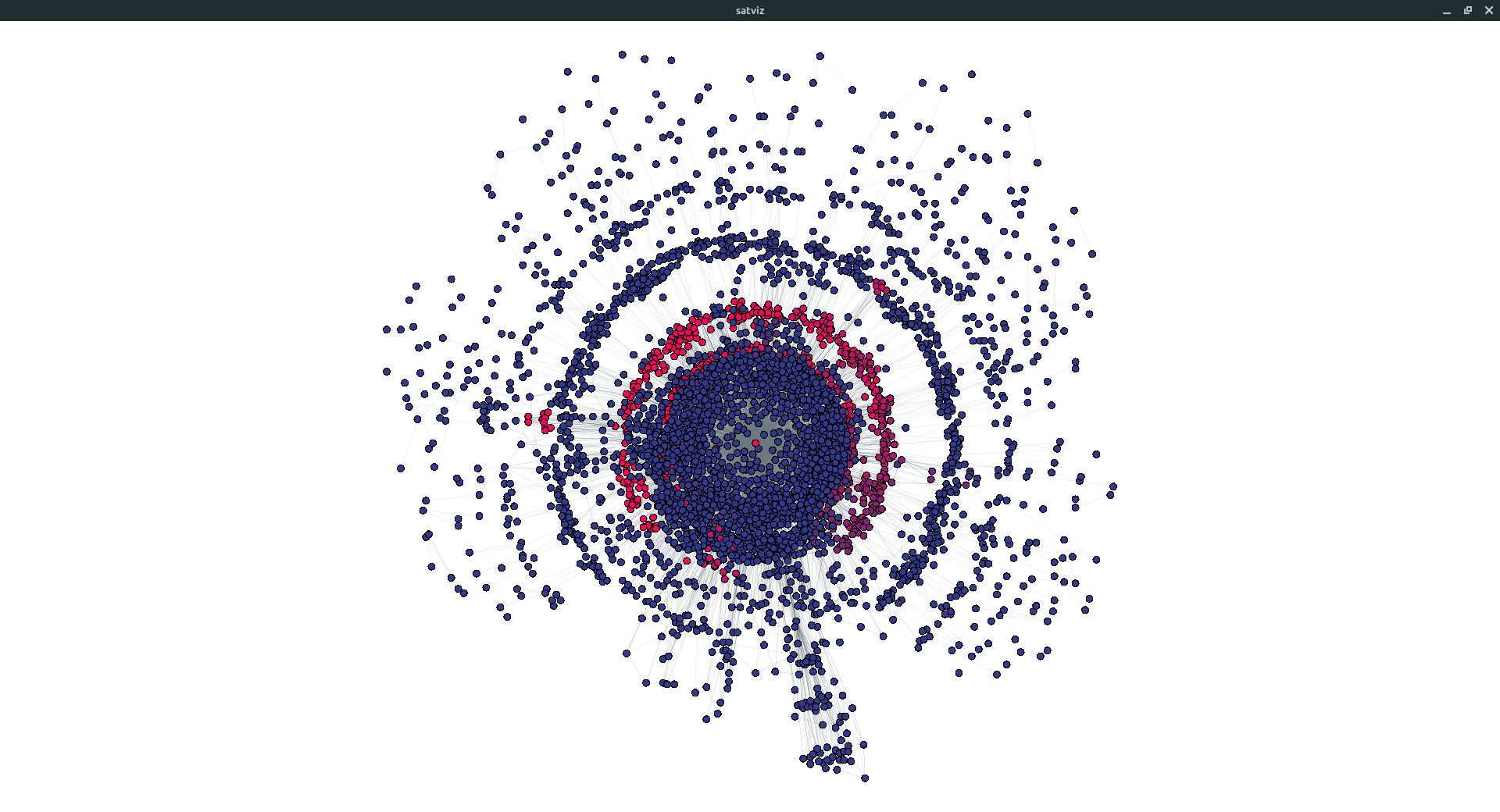}
\caption{Visualization of instance \textsf{AProVE09-15} (94663 variables) highlighting clauses learned by \textsf{Kissat} with no contraction (left) and 10 iterations of max-weight contractions (right).}
\label{fig:contraction}
\end{figure}

The size of SAT instances have a large impact on visualization performance. 
In initial experiments, the time to layout SAT instances with more than 50k variables reached a for usability critical time limit of one minute. 
Moreover, visualizations of larger SAT instances can become cluttered and their structure concealed as too many nodes compete for space. 
For these reasons, \satviz features a pre-processing step in which it runs a graph contraction algorithm which is based on the well-known label propagation algorithm~\cite{LABEL_PROPAGATION}.
The algorithm assigns labels to the nodes. Initially, each node has its own label. Then, the algorithm iterates over the nodes
in random order -- also called a \emph{round} -- and whenever a node is visited, it is assigned the label appearing most frequently in its neighborhood.
This is repeated for a fixed number of rounds or until none of the nodes changed its label in a round.
Subsequently, we collapse nodes with the same label into a single node and connect them to nodes corresponding to adjacent labels.
We repeat the clustering and contraction procedure recursively until the size of the graph becomes manageable for our graph drawing algorithm. 

Figure~\ref{fig:contraction} shows two visualizations of the SAT instance \textsf{AProVE09-15}\footnote{\url{https://gbd.iti.kit.edu/file/8a244542d09e20e9e8813ce7089c4135}} which has 94663 variables.
The left visualization represents the input instance, while the right visualization was obtained with our graph contraction algorithm.
As it can be seen, the graph contractions facilitate the analysis of large SAT instances
and can unveil hidden structures.

\subsection{Highlighting Areas of Interest}
\label{sec:heatmap}

\begin{figure}[t]
\centering
\includegraphics[width=.3\linewidth, trim={450 0 450 30px}, clip]{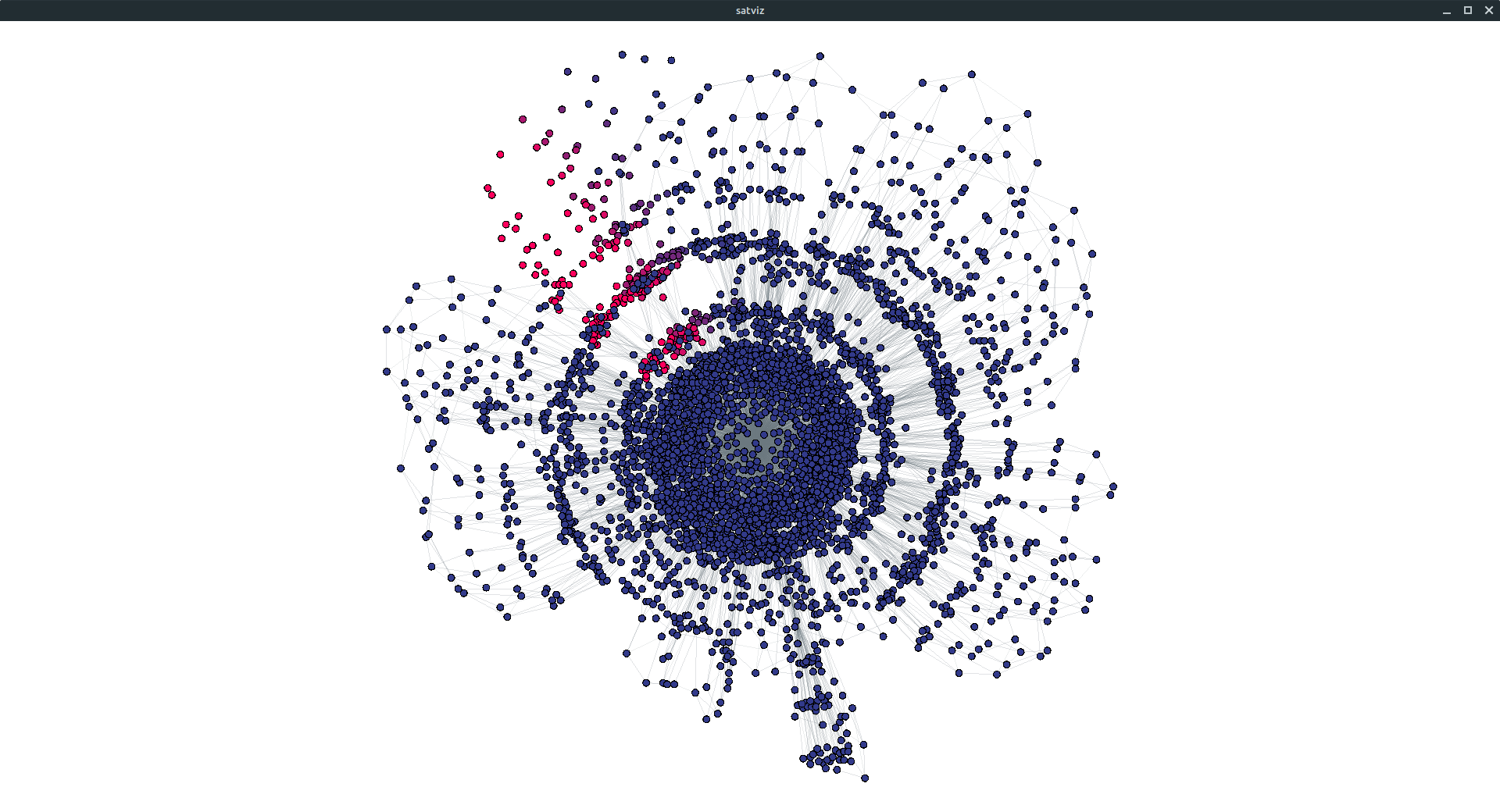}
\includegraphics[width=.3\linewidth, trim={450 0 450 30px}, clip]{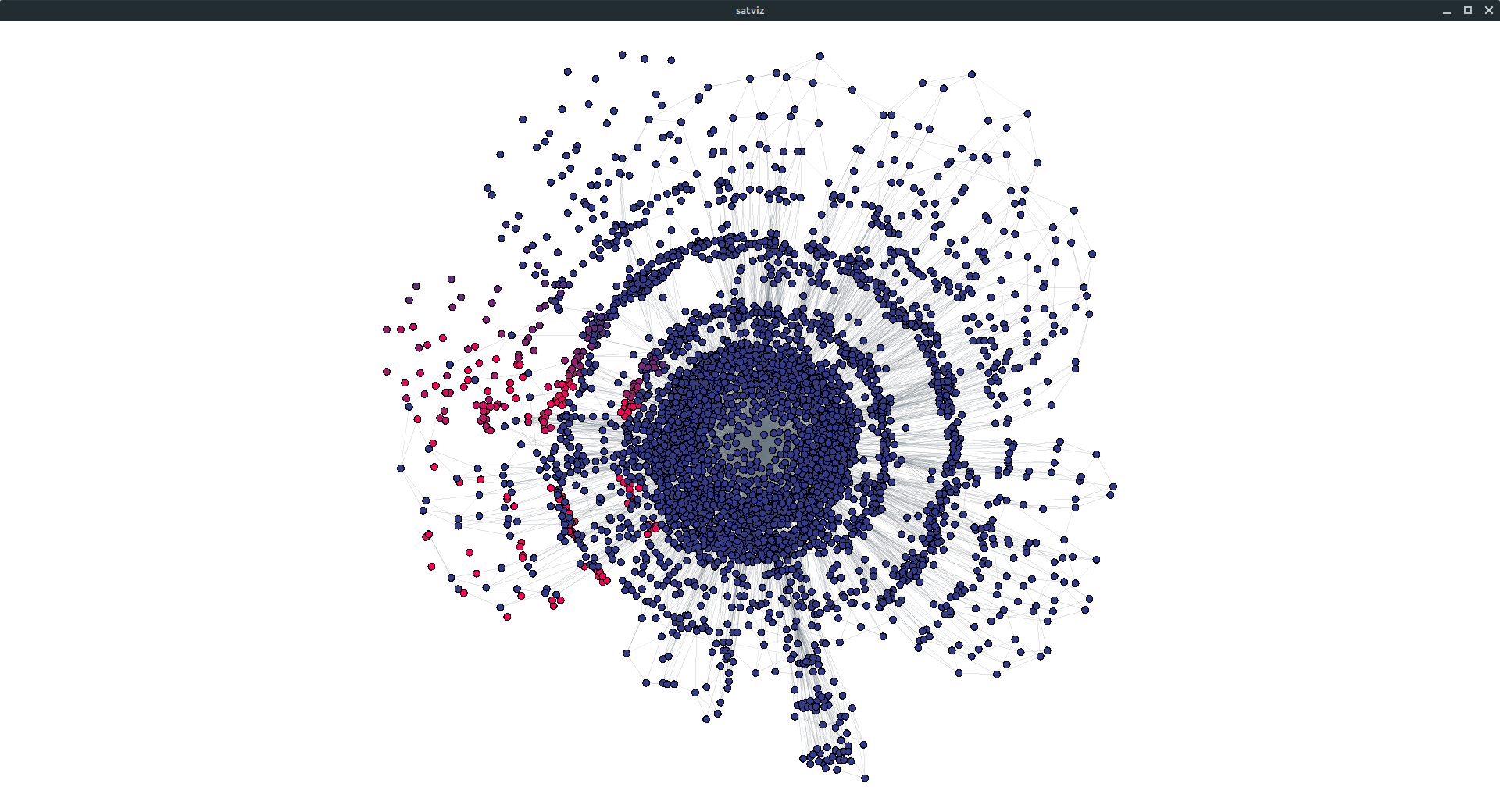}
\includegraphics[width=.3\linewidth, trim={450 0 450 30px}, clip]{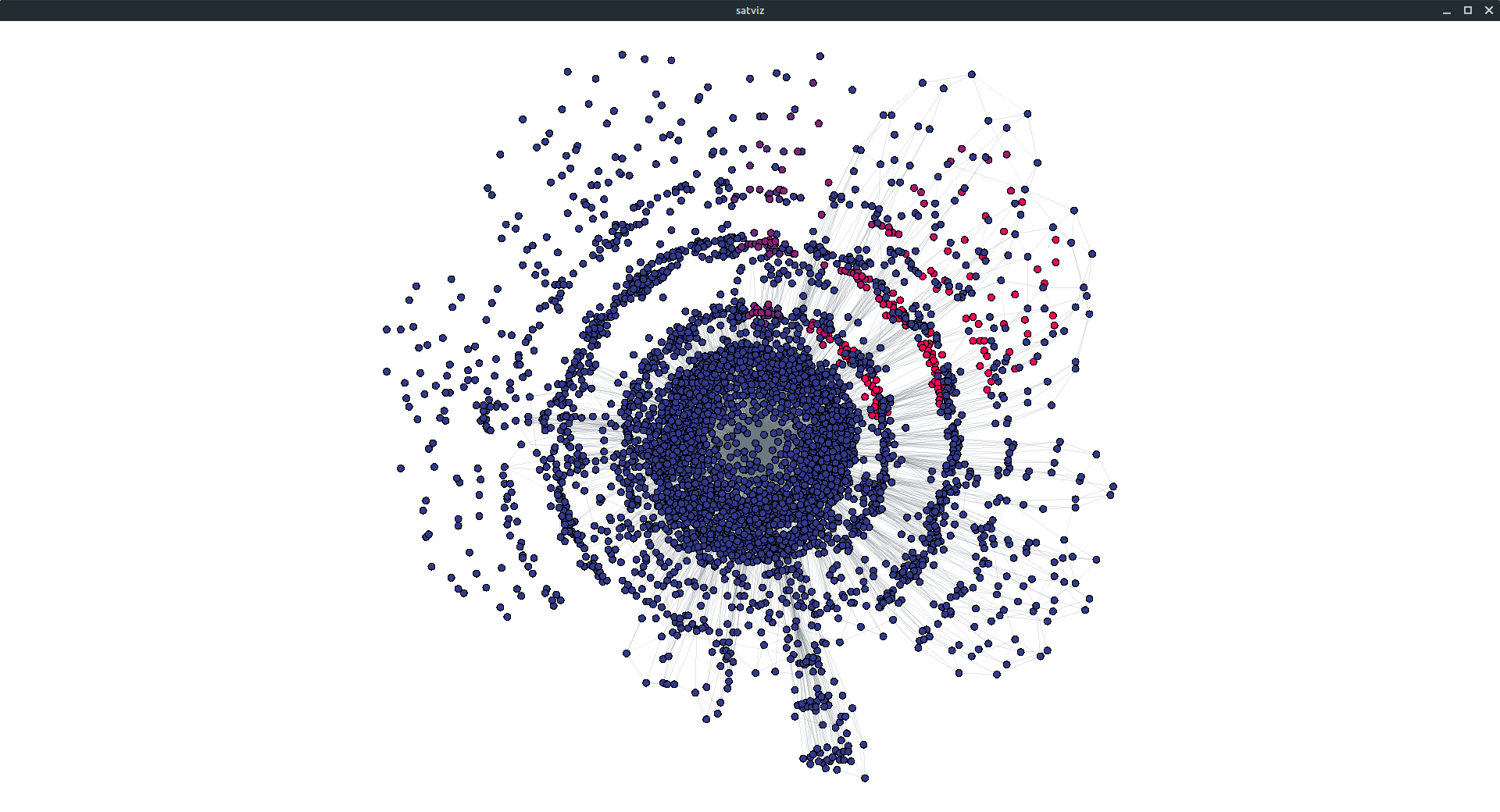}
\caption{Visualization of instance \textsf{AProVE09-15} highlighting clauses learned by \textsf{Kissat} -- with a heat-map frame width of 1000 clauses -- after 1000, 1690, and 3090 conflicts (from left to right).}
\label{fig:aprove-5000}
\end{figure}

\satviz uses configurable heat-map colors to highlight recently learned clauses.
We use a \emph{heat value} to determine the color for each node.
This value is determined by either (i) normalizing each variables' occurrence count for the last $k$ learned clauses or
(ii) assigning the maximum heat value to the most recently used variables and then reduce
the value to zero again within the next $k$ time steps (unless it appears again in a learned clause).
The parameter $k$ is configurable and assists in switching between coarse- and fine-grained proof analysis.
The color of node contractions is determined by the average heat value of the variables which it represents. 
\satviz also has a parameter for adjusting the speed of the animation.

Figure~\ref{fig:aprove-5000} shows three snapshots of a run of \textsf{Kissat}~\cite{kissat} on the instance \textsf{AProVE09-15} within the first 5000 learned clauses.
The heat-map highlights variables occurring in the $k=1000$ most recently learned clauses.
The displayed sequence of resolution steps appears in the animation like a windshield wiper.

\subsection{Evolution of a Proof via Relayouting}
\label{sec:relayout}

\begin{figure}[t]
\centering
\begin{adjustbox}{width=\textwidth}
\begin{tikzpicture}[line width=1.2pt, normalfg, node distance=5em]
\tikzset{default/.style={black, draw=normalbg, line width=20pt, fill=normalbg!20, minimum size=0.5cm, font=\relsize{1.5}\sffamily}}

\node[default, draw=white] (relayout) at (current page.east) {\includegraphics[width=50em, trim={450 0 450 30px}, clip]{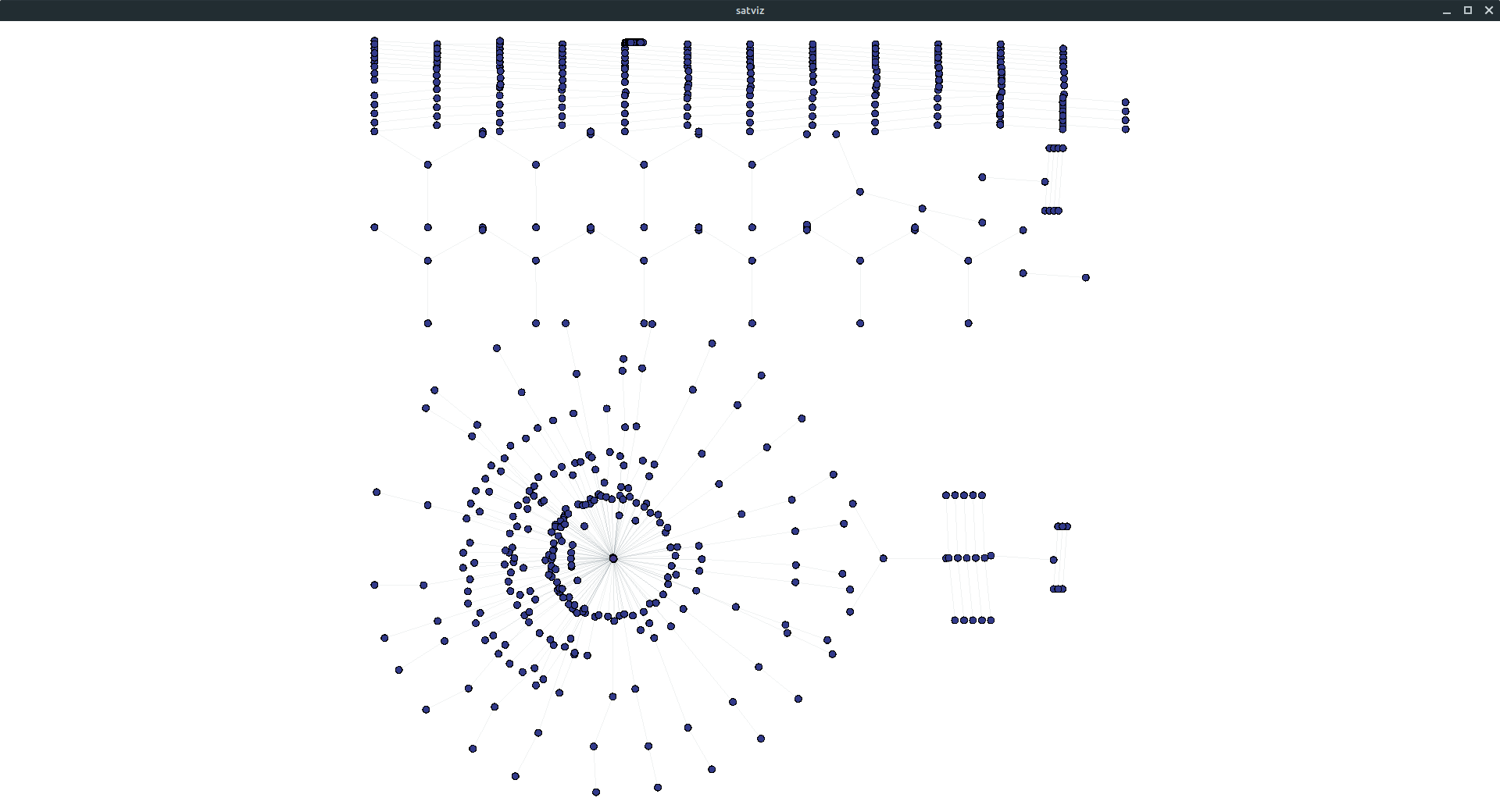}};
\node[default] (mag) at (current page.west) {\includegraphics[width=50em, trim={470 30 470 60px}, clip]{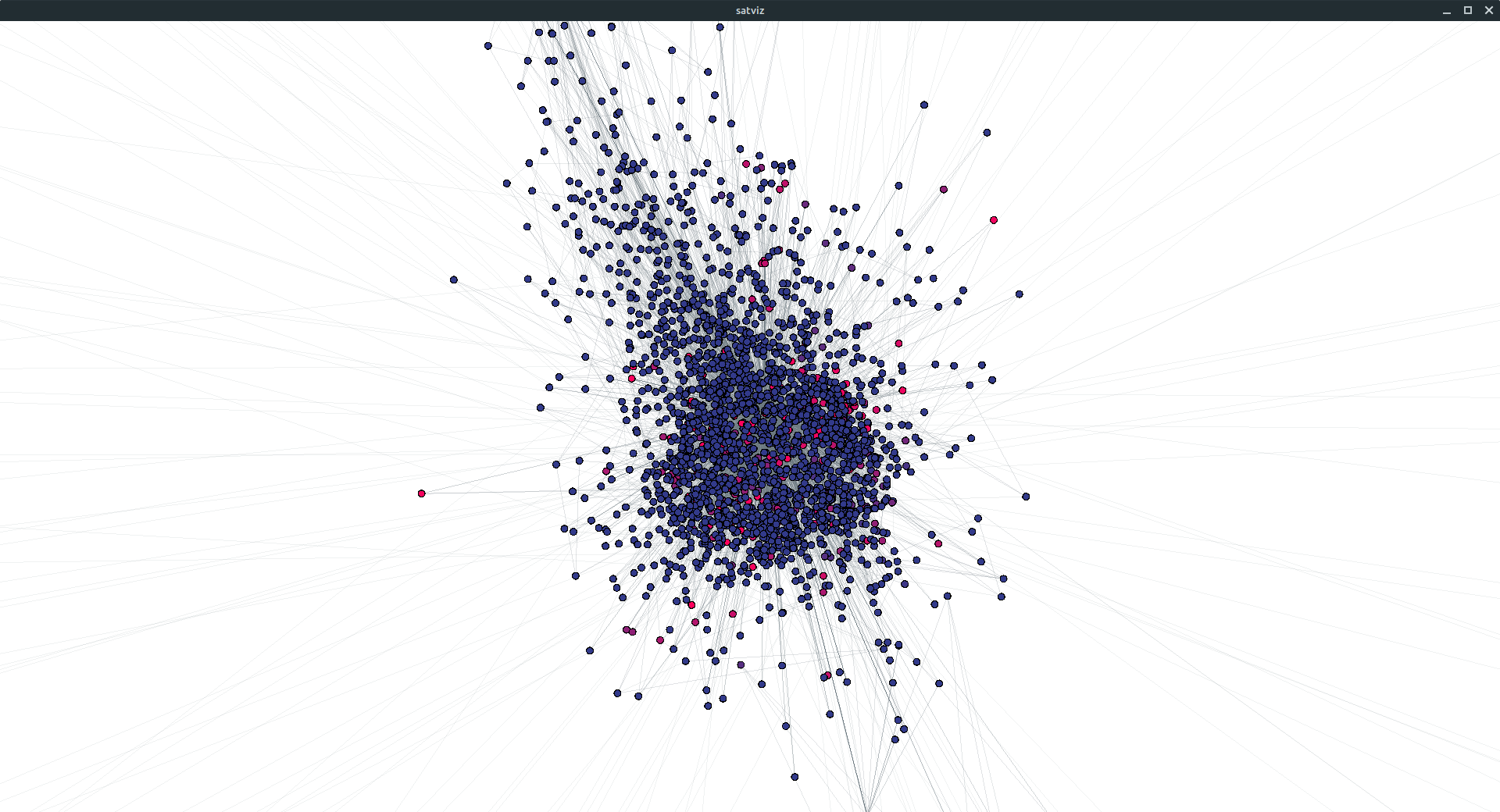}};

\coordinate (a) at ($ (mag.north east)+(6.7,-13.6) $);
\coordinate (b) at ($ (a)+(0,-1) $);
\coordinate (c) at ($ (b)+(1,0) $);
\coordinate (d) at ($ (a)+(1,0) $);

\draw[line width=5pt, draw=normalbg] (a) -- (b) -- (c) -- (d) -- (a);

\draw[line width=10pt, draw=normalbg] (mag.north east) -- (d);
\draw[line width=10pt, draw=normalbg] (mag.south east) -- (c);


\end{tikzpicture}
\end{adjustbox}
\caption{Visualization of instance \textsf{AProVE09-15} solved by \textsf{Kissat}. 
The right graph shows the instance after a relayout using adapted weights after 6000 learned or forgotten clauses.
The left graph shows the magnified center of the circular region of the right graph.}
\label{fig:aprove-6000}
\end{figure}
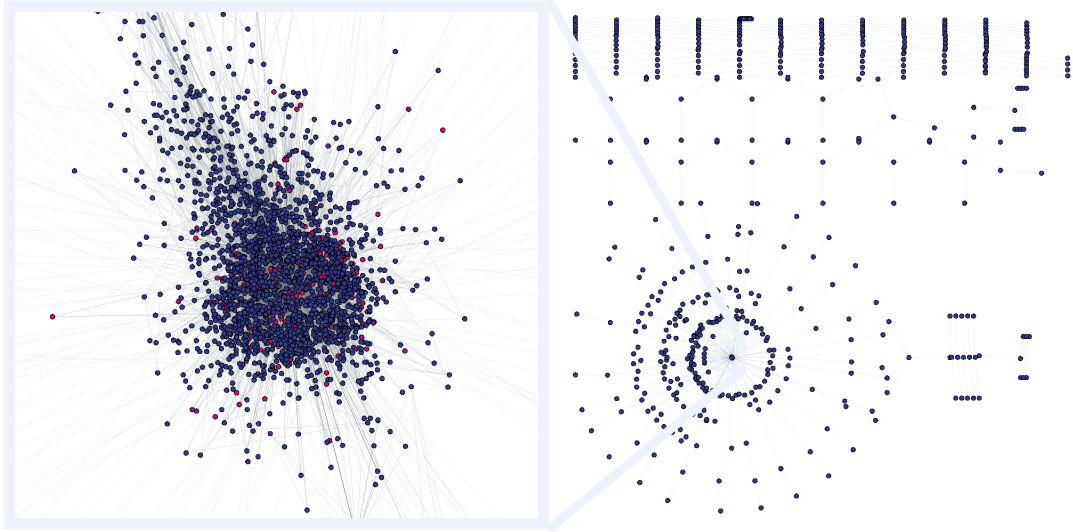

\satviz offers to pause the animation and relayout the instance with the weights adjusted by the learned and deleted clauses.
Figure~\ref{fig:aprove-6000} shows the structural changes in the \textsf{AProVE09-15} instance after 6000 clauses learned by \textsf{Kissat}.
We observe several variable connections become loose and degenerate to simple structures, which means that many original clauses become redundant and get deleted within the first 6000 conflicts. 
However, some variable connections get stronger, as the graph forms a densely connected community at the center of the circular structure of the relayouted instance in Figure~\ref{fig:aprove-6000}.

\subsection{Observing the Evolution of Formula Structure}

\begin{figure}[t]
\centering
\includegraphics[width=.461\linewidth, trim={460 30 460 30px}, clip, cfbox=normalbg 3pt 0pt]{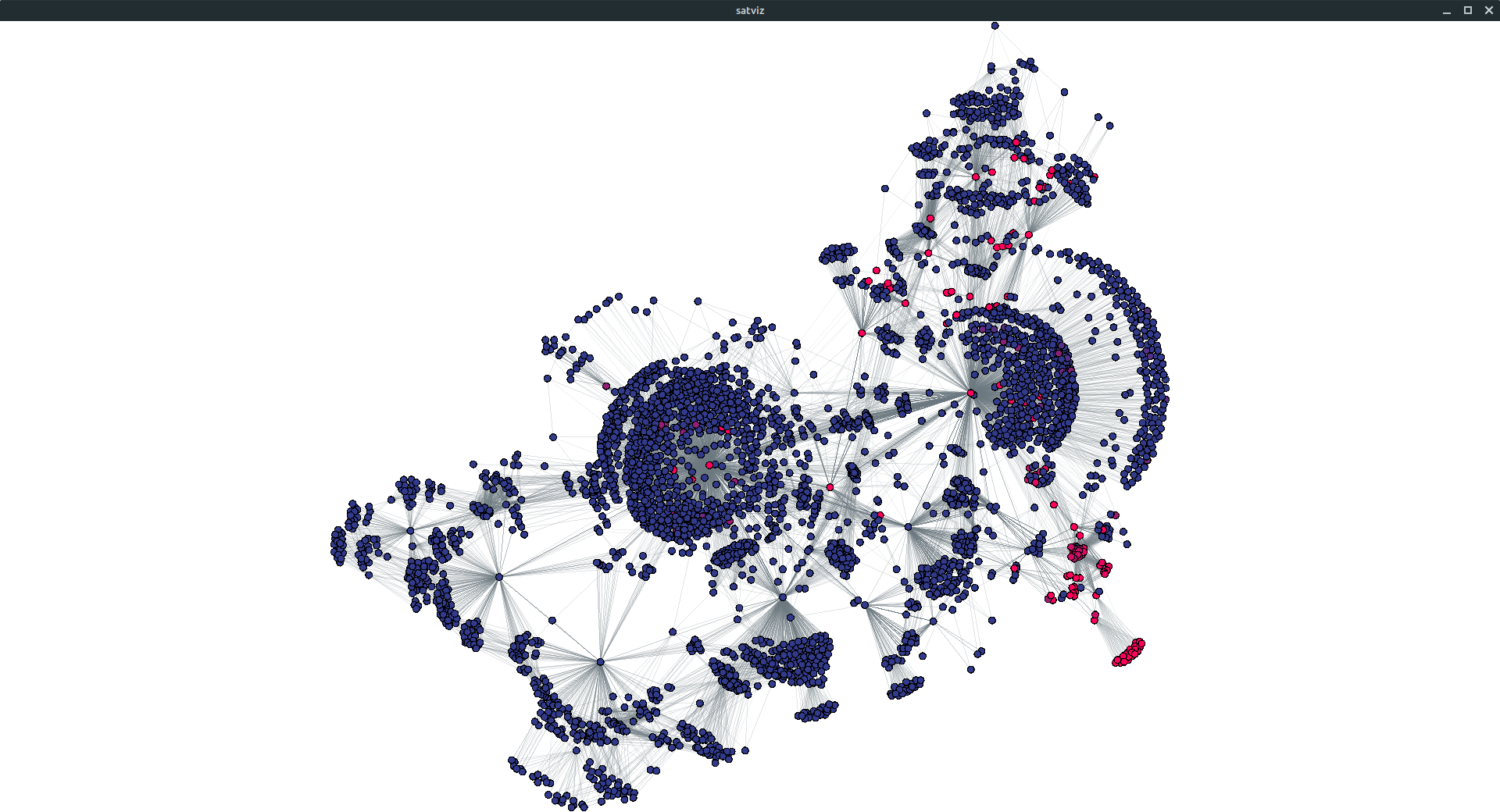}
\includegraphics[width=.461\linewidth, trim={460 30 460 30px}, clip, cfbox=normalbg 3pt 0pt]{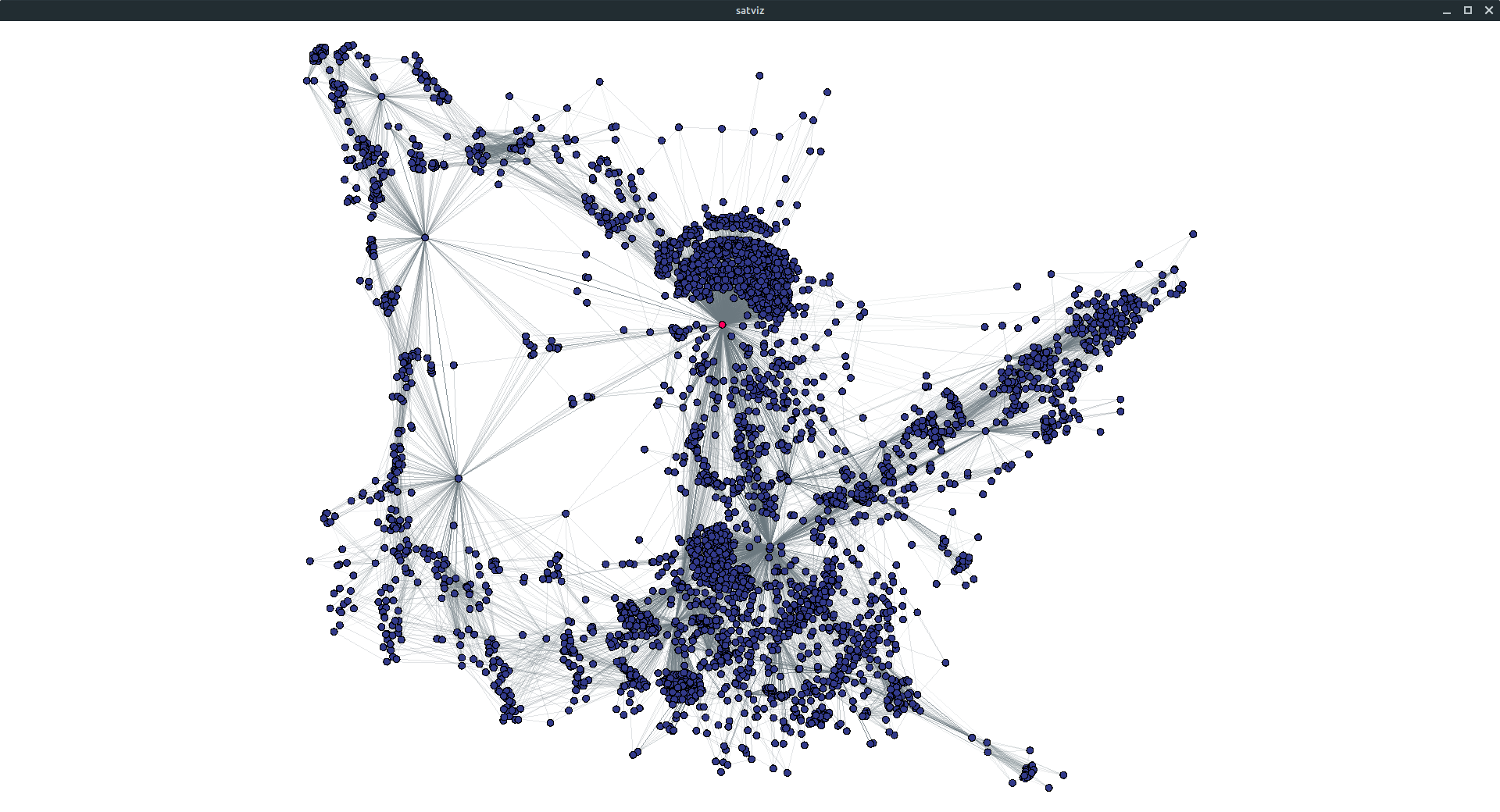}\\[.2em]
\includegraphics[width=.3\linewidth, trim={470 30 470 30px}, clip, cfbox=normalbg 3pt 0pt]{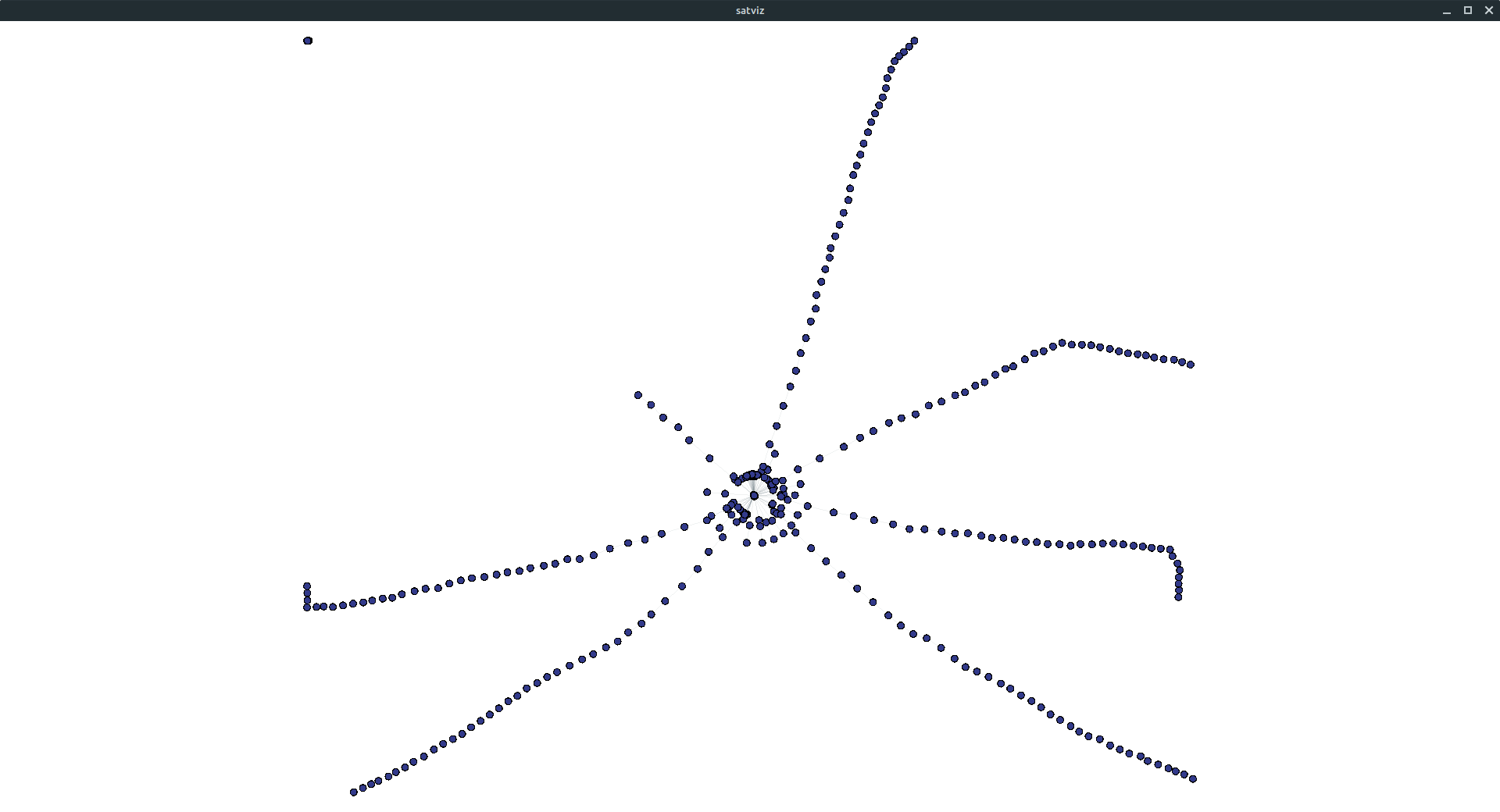}
\includegraphics[width=.3\linewidth, trim={470 30 470 30px}, clip, cfbox=normalbg 3pt 0pt]{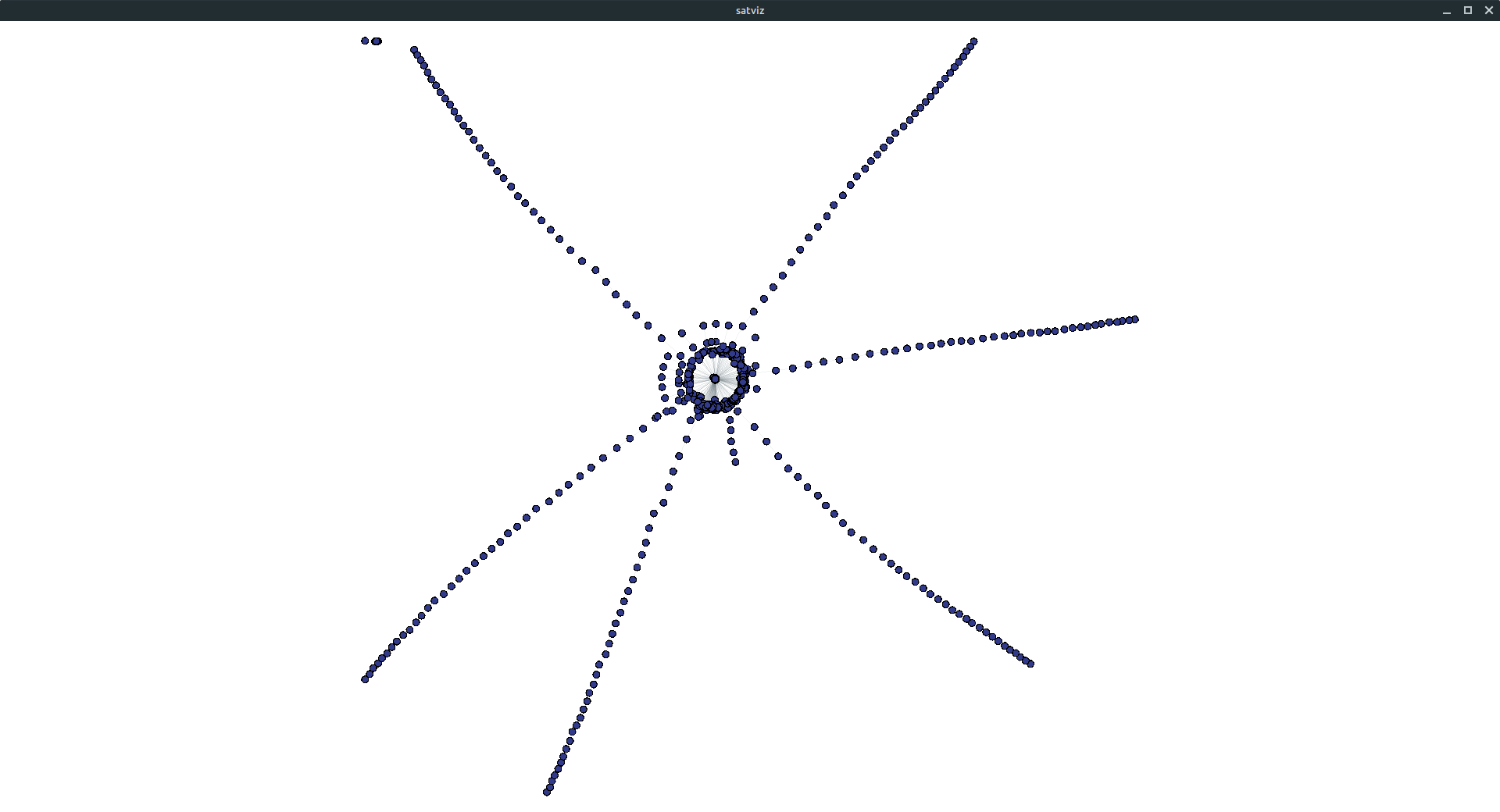}
\includegraphics[width=.3\linewidth, trim={470 30 470 30px}, clip, cfbox=normalbg 3pt 0pt]{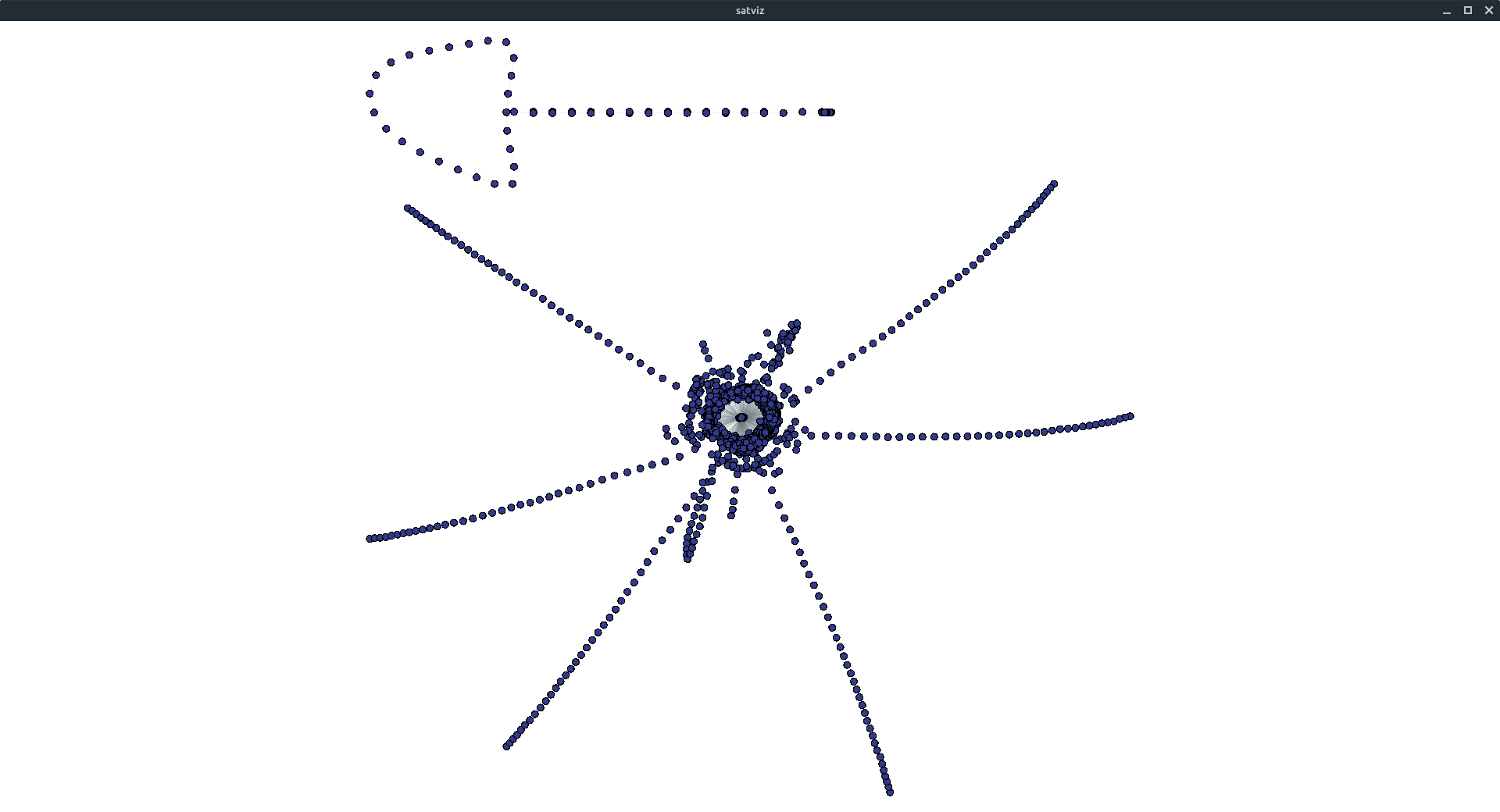}\\[.2em]
\includegraphics[width=.3\linewidth, trim={470 30 470 30px}, clip, cfbox=normalbg 3pt 0pt]{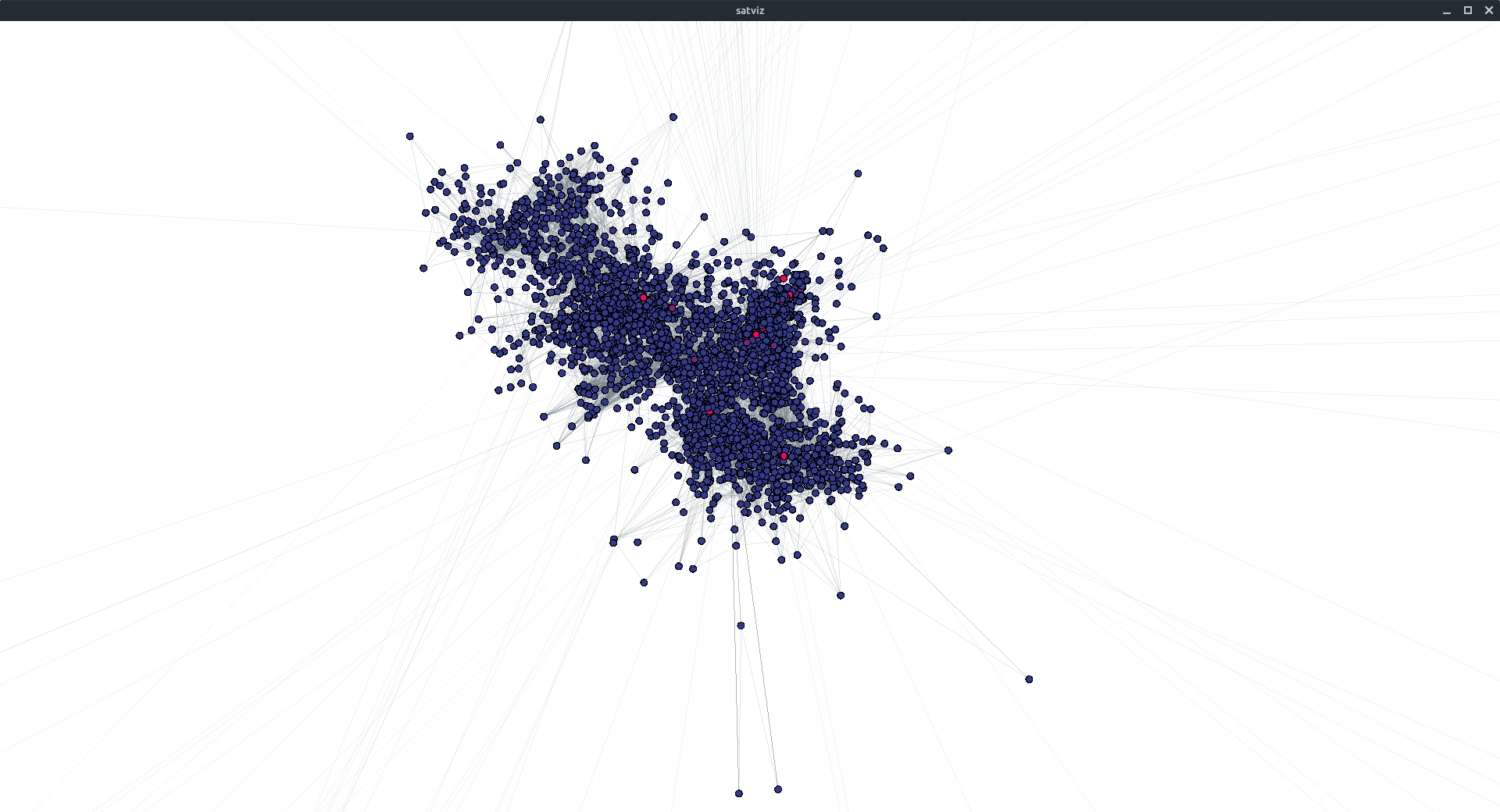}
\includegraphics[width=.3\linewidth, trim={470 30 470 30px}, clip, cfbox=normalbg 3pt 0pt]{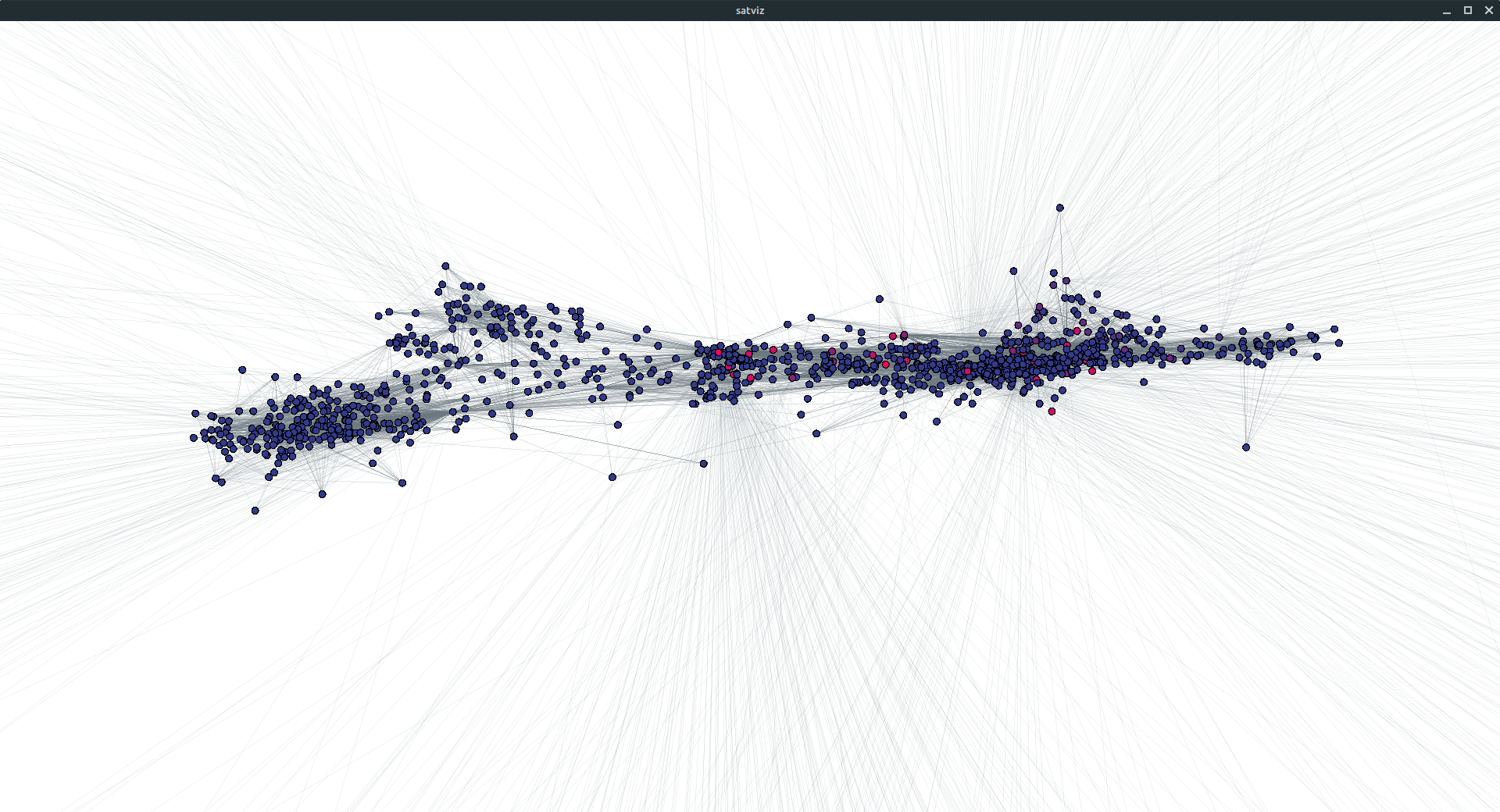}
\includegraphics[width=.3\linewidth, trim={470 30 470 30px}, clip, cfbox=normalbg 3pt 0pt]{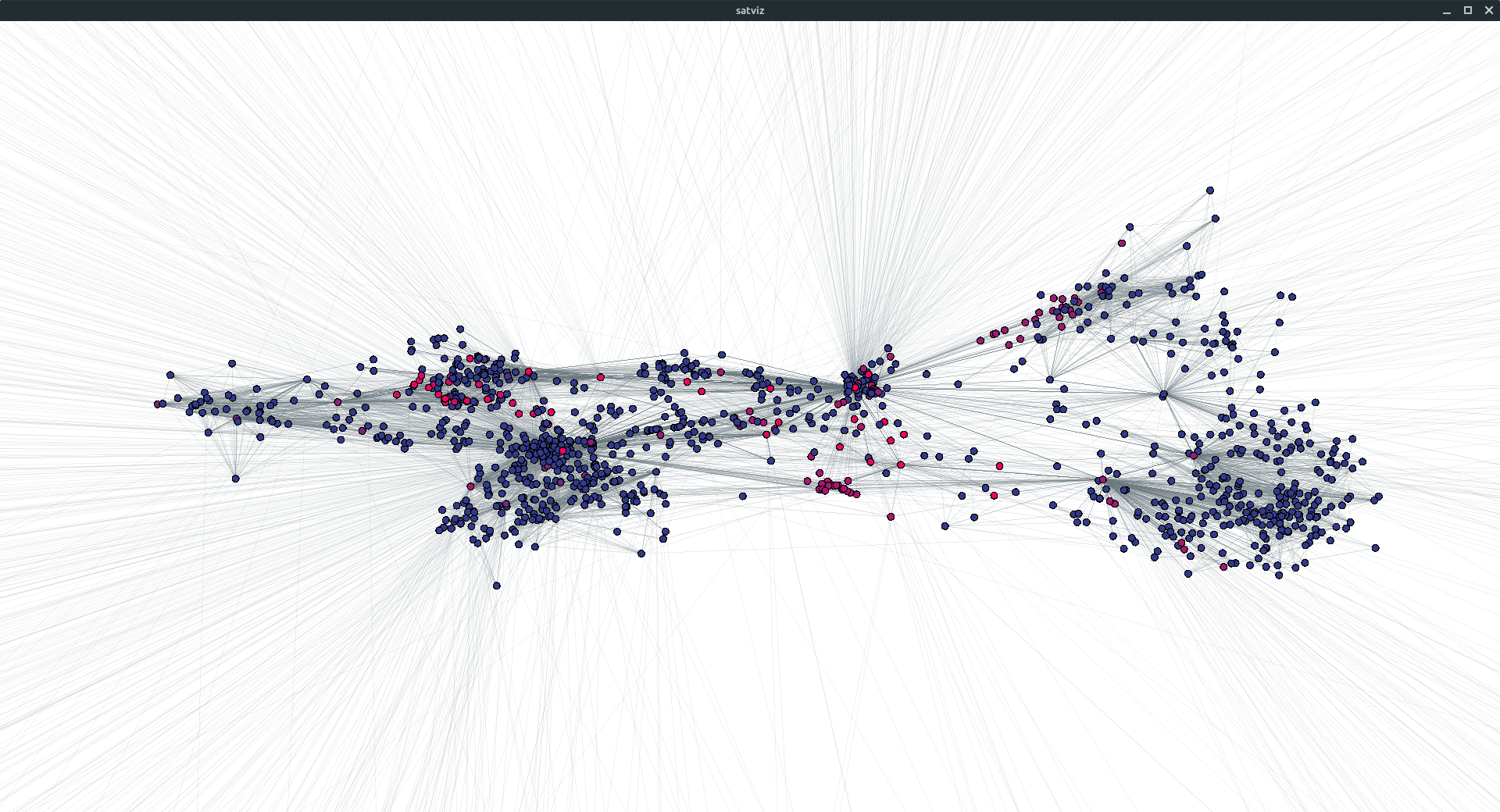}%
\caption{Visualization of instance \textsf{Newton.5.1.i.smt2-cvc4} solved by \textsf{Kissat}. The first row shows the instance in its original layout (left) and after 5000 learned or forgotten clauses (right). The second row shows the instance layout after one, two, and three million learned or forgotten clauses (from left to right). The third row shows the evolution of the magnified centers of the updated layouts after one, and three million learned or forgotten clauses and how it looks close to the of search at about 3.5 million learned or forgotten clauses (from left to right).}
\label{fig:evolution}
\end{figure}

Figure~\ref{fig:evolution} shows the evolution of the SAT instance \textsf{Newton.5.1.i.smt2-cvc4}\footnote{\url{https://gbd.iti.kit.edu/file/92b98ab8055b143e0283a215a70cf001}} as it was solved by \textsf{Kissat}. 
We observe that the original structure decays and a new structure emerges in which several nodes form circular shapes and tentacles. 
However, the resulting tentacles and circles are arranged around a densely connected core of nodes. 
This is due to several variables becoming weakly connected or even disconnected from the densely connected instance core. 
This densely connected core is where further searching and learning takes place, while nothing happens to the nodes that are now loosely connected. 
As learning progresses, more and more nodes detach from this core, so that it thins out and becomes flatter, while the inner circles around this core become more densely populated.

\section{Future Work and Conclusion}

Little research has gone into incremental weight updates in a force-directed layout algorithm.
We conducted initial experiments with real-time incremental adjustment of force-directed node placement.
However, it was problematic to regulate the extend of which node positions are allowed to change in such incremental updates.
Developing a stable incremental placement algorithm is a research project on its own.

Our future work will mainly focus on the consolidation and extension of \satviz in the following respects. 
\satviz should be able to process additional input about \emph{named equivalence classes of variables}. 
Such extra information can be displayed, e.g., by using additional node colors or when hovering nodes with the mouse. 
Equivalence classes of variables can also be used to specify preferences for the contraction algorithm or to increase the node proximity for such related variables with an adapted layout algorithm.
With the possibility to process and use classes of variables, \satviz can be even more helpful in analyzing large proofs and in identifying patterns and stages in proof generation. 

We also want to increase the number of available graph reduction methods and parameters to control them. 
As \satviz decouples clause producers via sockets, \satviz can also be extended to monitor clause exchange of parallel SAT solvers in the cloud. 

Visually observing the effects of the methods and heuristics that control clause learning and forgetting, or even clause sharing, can deepen our understanding of these methods and how they interact. 
The visualizations of \satviz can even leverage our understanding of the coarse structure of clausal proofs. 
\satviz can help identify and analyze arguments that consist of a large number of clauses. 
Recognizing patterns in clause proofs can help us better understand, compress, and explain them.

\bibliographystyle{plain}
\bibliography{main}

@inproceedings{Oh:2015:three-tier,
  author    = {Chanseok Oh},
  editor    = {Marijn Heule and
               Sean A. Weaver},
  title     = {Between {SAT} and {UNSAT:} The Fundamental Difference in {CDCL} {SAT}},
  booktitle = {Theory and Applications of Satisfiability Testing - {SAT} 2015 - 18th
               International Conference, Austin, TX, USA, September 24-27, 2015,
               Proceedings},
  series    = {Lecture Notes in Computer Science},
  volume    = {9340},
  pages     = {307--323},
  publisher = {Springer},
  year      = {2015}
}

@inproceedings{Feng:2020:AllUIP,
  author    = {Nick Feng and
               Fahiem Bacchus},
  editor    = {Luca Pulina and
               Martina Seidl},
  title     = {Clause Size Reduction with all-UIP Learning},
  booktitle = {Theory and Applications of Satisfiability Testing - {SAT} 2020 - 23rd
               International Conference, Alghero, Italy, July 3-10, 2020, Proceedings},
  series    = {Lecture Notes in Computer Science},
  volume    = {12178},
  pages     = {28--45},
  publisher = {Springer},
  year      = {2020}
}

@article{Sinz:2007:DPvis,
  author    = {Carsten Sinz},
  title     = {Visualizing {SAT} Instances and Runs of the {DPLL} Algorithm},
  journal   = {J. Autom. Reason.},
  volume    = {39},
  number    = {2},
  pages     = {219--243},
  year      = {2007},
  url       = {https://doi.org/10.1007/s10817-007-9074-1}
}

@book{Knuth:2015:TaocpSat,
  title={The art of computer programming, Volume 4, Fascicle 6: Satisfiability},
  author={Knuth, Donald E},
  year={2015},
  publisher={Addison-Wesley Professional}
}

@inproceedings{Newsham:2015:SATGraf,
  author    = {Zack Newsham and
               William Lindsay and
               Vijay Ganesh and
               Jia Hui Liang and
               Sebastian Fischmeister and
               Krzysztof Czarnecki},
  editor    = {Marijn Heule and
               Sean A. Weaver},
  title     = {SATGraf: Visualizing the Evolution of {SAT} Formula Structure in Solvers},
  booktitle = {Theory and Applications of Satisfiability Testing - {SAT} 2015 - 18th
               International Conference, Austin, TX, USA, September 24-27, 2015,
               Proceedings},
  series    = {Lecture Notes in Computer Science},
  volume    = {9340},
  pages     = {62--70},
  publisher = {Springer},
  year      = {2015}
}

@inproceedings{Buning:2020:QPRverify,
  author    = {Marko Kleine B{\"{u}}ning and
               Carsten Sinz and
               David Farag{\'{o}}},
  editor    = {Maria Christakis and
               Nadia Polikarpova and
               Parasara Sridhar Duggirala and
               Peter Schrammel},
  title     = {{QPR} Verify: {A} Static Analysis Tool for Embedded Software Based
               on Bounded Model Checking},
  booktitle = {Software Verification - 12th International Conference, {VSTTE} 2020,
               and 13th International Workshop, {NSV} 2020, Los Angeles, CA, USA,
               July 20-21, 2020, Revised Selected Papers},
  series    = {Lecture Notes in Computer Science},
  volume    = {12549},
  pages     = {21--32},
  publisher = {Springer},
  year      = {2020},
  url       = {https://doi.org/10.1007/978-3-030-63618-0\_2}
}

@inproceedings{Kaufmann:2021:AMulet,
  author    = {Daniela Kaufmann and
               Armin Biere},
  editor    = {Jan Friso Groote and
               Kim Guldstrand Larsen},
  title     = {AMulet 2.0 for Verifying Multiplier Circuits},
  booktitle = {Tools and Algorithms for the Construction and Analysis of Systems
               - 27th International Conference, {TACAS} 2021, Held as Part of the
               European Joint Conferences on Theory and Practice of Software, {ETAPS}
               2021, Luxembourg City, Luxembourg, March 27 - April 1, 2021, Proceedings,
               Part {II}},
  series    = {Lecture Notes in Computer Science},
  volume    = {12652},
  pages     = {357--364},
  publisher = {Springer},
  year      = {2021},
  url       = {https://doi.org/10.1007/978-3-030-72013-1\_19}
}

@inproceedings{Werner:2021:ValDom,
  author    = {Johannes Werner and
               Tom{\'{a}}s Balyo and
               Ashlin Iser and
               Michael Klein},
  editor    = {Michel Aldanondo and
               Andreas A. Falkner and
               Alexander Felfernig and
               Martin Stettinger},
  title     = {Fast Approximate Calculation of Valid Domains in a Satisfiability-based
               Product Configurator},
  booktitle = {Proceedings of the 23rd International Configuration Workshop (CWS/ConfWS
               2021), Vienna, Austria, 16-17 September, 2021},
  series    = {{CEUR} Workshop Proceedings},
  volume    = {2945},
  pages     = {24--32},
  publisher = {CEUR-WS.org},
  year      = {2021},
  url       = {http://ceur-ws.org/Vol-2945/23-JW-ConfWS21\_paper\_15.pdf}
}

@article{Schreiber:2021:Lilotane,
  author    = {Dominik Schreiber},
  title     = {Lilotane: {A} Lifted SAT-based Approach to Hierarchical Planning},
  journal   = {J. Artif. Intell. Res.},
  volume    = {70},
  pages     = {1117--1181},
  year      = {2021},
  url       = {https://doi.org/10.1613/jair.1.12520}
}

@inproceedings{Heule:2018:Schur,
	author    = {Marijn J. H. Heule},
	title     = {Schur Number Five},
	booktitle = {Proc. AAAI},
	location  = {New Orleans, Louisiana, USA},
	pages     = {6598--6606},
	year      = {2018},
	url       = {https://ojs.aaai.org/index.php/AAAI/article/view/12209}
}

@inproceedings{Heule:2016:Pyth,
	author    = {Marijn J. H. Heule and
	Oliver Kullmann and
	Victor W. Marek},
	title     = {Solving and Verifying the Boolean Pythagorean Triples Problem via Cube-and-Conquer},
	booktitle = {Proc. SAT},
	location  = {Bordeaux, France},
	pages     = {228--245},
	year      = {2016},
	doi       = {10.1007/978-3-319-40970-2_15}
}

@inproceedings{kissat,
  author    = {Armin Biere and Katalin Fazekas and Mathias Fleury and Maximillian Heisinger},
  title     = {{CaDiCaL}, {Kissat}, {Paracooba}, {Plingeling} and {Treengeling} Entering the {SAT Competition 2020}},
  pages     = {51--53},
  editor    = {Tomas Balyo and Nils Froleyks and Marijn Heule and Ashlin Iser and Matti J{\"a}rvisalo and Martin Suda},
  booktitle = {Proc.~of {SAT Competition} 2020 -- Solver and Benchmark Descriptions},
  volume    = {B-2020-1},
  series    = {Department of Computer Science Report Series B},
  publisher = {University of Helsinki},
  year      = 2020,
}

@article{drat-trim,
  author    = {Marijn J. H. Heule},
  title     = {The {DRAT} format and DRAT-trim checker},
  journal   = {CoRR},
  volume    = {abs/1610.06229},
  year      = {2016},
  url       = {http://arxiv.org/abs/1610.06229}
}

@book{sat-handbook,
  editor    = {Armin Biere and
               Marijn Heule and
               Hans van Maaren and
               Toby Walsh},
  title     = {Handbook of Satisfiability - Second Edition},
  series    = {Frontiers in Artificial Intelligence and Applications},
  volume    = {336},
  publisher = {{IOS} Press},
  year      = {2021}
}

@article{ipasir,
  author    = {Tom{\'{a}}s Balyo and
               Armin Biere and
               Ashlin Iser and
               Carsten Sinz},
  title     = {{SAT} Race 2015},
  journal   = {Artif. Intell.},
  volume    = {241},
  pages     = {45--65},
  year      = {2016}
}

@inproceedings{minisat,
  author    = {Niklas E{\'{e}}n and
               Niklas S{\"{o}}rensson},
  editor    = {Enrico Giunchiglia and
               Armando Tacchella},
  title     = {An Extensible SAT-solver},
  booktitle = {Theory and Applications of Satisfiability Testing, 6th International
               Conference, {SAT} 2003. Santa Margherita Ligure, Italy, May 5-8, 2003
               Selected Revised Papers},
  series    = {Lecture Notes in Computer Science},
  volume    = {2919},
  pages     = {502--518},
  year      = {2003}
}

@inproceedings{Orbe:2012:iSAT,
  author    = {Ezequiel Orbe and
               Carlos Areces and
               Gabriel G. Infante L{\'{o}}pez},
  editor    = {Nikolaj S. Bj{\o}rner and
               Andrei Voronkov},
  title     = {iSat: Structure Visualization for {SAT} Problems},
  booktitle = {Logic for Programming, Artificial Intelligence, and Reasoning - 18th
               International Conference, LPAR-18, M{\'{e}}rida, Venezuela, March
               11-15, 2012. Proceedings},
  series    = {Lecture Notes in Computer Science},
  volume    = {7180},
  pages     = {335--342},
  publisher = {Springer},
  year      = {2012}
}

@article{Chimani:2013:OGDF,
  title={The Open Graph Drawing Framework (OGDF).},
  author={Chimani, Markus and Gutwenger, Carsten and J{\"u}nger, Michael and Klau, Gunnar W and Klein, Karsten and Mutzel, Petra},
  journal={Handbook of graph drawing and visualization},
  volume={2011},
  pages={543--569},
  year={2013}
}

@incollection{BIPARTITE-GRAPH,
  author      = {Hu, T. C. and Moerder, K.},
  editor      = {Hu, T.C. and Kuh, E.S.},
  title       = {{Multiterminal Flows in a Hypergraph}},
  booktitle   = {VLSI Circuit Layout: Theory and Design},
  chapter     = {3},
  pages       = {87--93},
  publisher   = IEEE,
  year        = {1985}
}

@inproceedings{BIPARTITE-GRAPH-2,
  author    = {Daniel G. Schweikert and
               Brian W. Kernighan},
  title     = {{A Proper Model for the Partitioning of Electrical Circuits}},
  booktitle = DAC72,
  pages     = {57--62},
  year      = {1972},
  month     = {6}
}

@article{LABEL_PROPAGATION,
  title={{Near Linear Time Algorithm to Detect Community Structures in Large-Scale Networks}},
  author={Raghavan, Usha Nandini and Albert, R{\'e}ka and Kumara, Soundar},
  journal={Physical Review E},
  volume={76},
  number={3},
  pages={036106},
  year={2007},
  publisher={APS},
  doi={10.1103/PhysRevE.76.036106}
}

\end{document}